\pdfoutput=1
\documentclass[sigconf, nonacm]{acmart}

\setcopyright{acmcopyright}
\copyrightyear{2024}
\acmYear{2024}
\acmDOI{10.1145/3649329.3656521}
\acmISBN{}

\acmConference[DAC '24]{61st ACM/IEEE Design Automation Conference}{June 23--27,
  2024}{San Francisco, CA}

\usepackage{algorithm}
\usepackage{algorithmic}

\newtheorem{theorem}{Theorem}
\newtheorem{definition}{Definition}[section]
\newtheorem{lemma}[theorem]{Lemma}

\begin{document}

\title[SGM-PINN]{SGM-PINN: Sampling Graphical Models for Faster
Training of Physics-Informed Neural Networks}

\author{John Anticev}
\email{janticev@stevens.edu}
\affiliation{%
  \institution{Stevens Institute of Technology}
  \streetaddress{1 Castle Point}
  \city{Hoboken}
  \state{New Jersey}
  \country{USA}
  \postcode{07030}
}

\author{Ali Aghdaei}
\email{aaghdae1@stevens.edu}
\affiliation{%
  \institution{Stevens Institute of Technology}
  \streetaddress{1 Castle Point}
  \city{Hoboken}
  \state{New Jersey}
  \country{USA}
  \postcode{07030}
}

\author{Wuxinlin Cheng}
\email{wcheng7@stevens.edu }
\affiliation{%
  \institution{Stevens Institute of Technology}
  \streetaddress{1 Castle Point}
  \city{Hoboken}
  \state{New Jersey}
  \country{USA}
  \postcode{07030}
}

\author{Zhuo Feng}
\email{zfeng12@stevens.edu }
\affiliation{%
  \institution{Stevens Institute of Technology}
  \streetaddress{1 Castle Point}
  \city{Hoboken}
  \state{New Jersey}
  \country{USA}
  \postcode{07030}
}


\begin{abstract}
    SGM-PINN is a graph-based importance sampling framework to improve the training efficacy of Physics-Informed Neural Networks (PINNs) on parameterized problems. By applying a graph decomposition scheme to an undirected Probabilistic Graphical Model (PGM) built from the training dataset, our method generates node clusters encoding conditional dependence between training samples. Biasing sampling towards more important clusters allows smaller mini-batches and training datasets, improving training speed and accuracy. We additionally fuse an efficient robustness metric with residual losses to determine regions requiring additional sampling. Experiments demonstrate the advantages of the proposed framework, achieving $3\times$ faster convergence compared to prior state-of-the-art sampling methods.
\end{abstract}

\maketitle
\section{Introduction}
Over the past few decades, partial differential equation (PDE) solvers have been playing important roles in numerous compute-intensive computer-aided design (CAD) tasks, including circuit/device simulations \citep{fichtner1983semiconductor},   chip thermal analysis \citep{li2004efficient}, computational electromagnetics (CEM) \citep{inan2011numerical}, computational fluid dynamics analysis (CFD) \citep{tu2018computational}. Traditional numerical PDE solvers like finite difference methods (FDM) and finite element analysis (FEA),  tackle the problem through spatial discretization. However, finer discretization for better simulation accuracy slows down performance due to the super-linear complexity of existing sparse matrix solvers. These classic PDE algorithms often require finer grids, which can be very expensive in terms of both time and hardware resources, potentially taking days on powerful computing clusters.
 
An emerging class of physics-informed neural-network (PINN) based methods use automatic differentiation to estimate the solution of PDEs by training a neural network (NN) to minimize the residuals of the governing equations, initial/boundary conditions, and measurement data over a set of collocation points defined on a domain tailored to each problem \citep{hennigh2021nvidia,raissi2019physics}. These PINN-based solvers also address problems not easily solvable by traditional methods, such as inverse or data assimilation problems, as well as real-time simulation problems \cite{raissi2019physics,xu2019neuralinverse}. Additionally, and explored in this study, they can simultaneously learn the solution to a problem across parameterized geometries~\citep{hennigh2021nvidia,raissi2019physics}; such as varying the design, materials, or size of the fin stack of a heatsink. As a result, the PINN-based PDE solvers can achieve orders of magnitude faster performance than conventional solvers when many variations of a design are desired. For example, a trained parameterized PINN may be used as an ultra-fast surrogate model for another gradient-based optimization method which inferences the neural network for approximate solutions across thousands of variations of a design \citep{xue2020amortized}.  

While PINNs have demonstrated exceptional performance across a variety of problems related to PDE solvers, there are many cases in which PINNs fail to converge, or do so to a trivial or inaccurate solution \cite{hao2023physicsinformedsurveyproblems,faroughi2023physicsguidedsurveyproblems,wang2020pinnsNTK}. Although better convergence can be sometimes be realized with higher-density point clouds and larger batch sizes, this imposes memory constraints that tax the GPU hardware most commonly used to accelerate deep learning workloads. This leads to long training times that require the highest-end hardware, ultimately mirroring the issues with traditional PDE solvers exhibit in terms of scalability with model size and complexity. 

Among the proposed solutions to improve the convergence and efficiency of PINN training are residual-based adaptive refinement (RAR) methods \cite{DeepXDE} (where additional samples are added to regions with high PDE residuals) and importance sampling (IS) methods \cite{Nvidia-IS} (where each mini-batch of samples is selected via a distribution proportional to the loss value at each sample in a dense dataset). These are implemented in currently available PINN solvers such as DeepXDE~\cite{DeepXDE} and Modulus Sym~\citep{hennigh2021nvidia}. Both methods suffer from high computational complexity and overhead associated with frequently calculating the residuals for every sample in a dense set, and can lead to poor retention of the solution on low-residual parts of the domain as training continues \cite{daw2023mitigatingR3}.

To address the limitations of existing PINN-based PDE solvers, we introduce an importance-sampling framework (SGM-PINN) to exploit the conditional dependence among the training samples by leveraging probabilistic graphical models (PGMs). Instead of training a PINN model using the full set of data samples (point cloud) during the training process that may involve millions of stochastic gradient descent (SGD) iterations, SGM-PINN adaptively forms smaller epochs by selecting the most representative data samples with a graph-based importance sampling strategy. Our preliminary results show that SGM-PINN can achieve up to $3\times$ faster convergence for training (parameterized) PINNs on several large-scale PDE problems.   Our contributions include:
\begin{enumerate}
    \item  We present SGM-PINN, a scalable graph-based IS framework based on spectral graph clustering, to improve PINN training. The key component in SGM-PINN is a low-\-resistance\--diameter (LRD) decomposition scheme for partitioning a PGM (graph) into strongly coupled sample clusters that can be used to reduce the computational overhead of estimating the importance score across the input data.
\item We show the proposed method allows for retaining good solution quality while reducing the batch size and the number of sample points to speed up training.
\item We demonstrate the addition of a spectral stability metric \cite{cheng:icml21} to incorporate gradient information of the loss with respect to input data to augment sample importance scoring, in order to improve parameterized PINN training.
\end{enumerate}

The rest of the paper is organized as follows: 
Section \ref{sec:background} provides a brief introduction to the theoretical foundation of PINNs and IS methods;  
Section \ref{sec:overview} gives an overview of the proposed SGM-PINN framework; 
Section \ref{sec:results} demonstrates experimental results to evaluate the performance of SGM-PINN, and 
Section \ref{sec:conclusion} that concludes this work.

\section{Background}\label{sec:background}
  \subsection{Theoretical background of PINNs.}
 Consider the following general form of a PDE equation under a set of boundary condition (BC) and initial condition (IC) constraints \citep{hennigh2021nvidia}:
\begin{equation}\label{eqn:pde}
\begin{aligned}
   & \mathcal{F}_i[u](\textbf{x}) = f_i(\textbf{x}), \,\,\,   \forall i\in \{1, \cdots,  N_\mathcal{F}\}, \textbf{x} \in \mathcal{D} \\
    &\mathcal{C}_j[u](\textbf{x}) = g_j(\textbf{x}), \,\,\,   \forall j\in \{1, \cdots, N_\mathcal{C}\}, \textbf{x} \in \partial \mathcal{D},
\end{aligned}
\end{equation}
where $\mathcal{F}_i$ is the general differential operator, $N_\mathcal{F}$ ($N_\mathcal{C}$) denotes the number of BCs (ICs), $\textbf{x}$ is the set of independent variables defined over a bounded continuous domain $\mathcal{D} \subseteq \mathbb{R}^d$, $u(\textbf{x})$ is the solution to the PDE, $\mathcal{C}_j$ is the constraint operator---such as the differential, linear, or nonlinear terms representing the boundary and initial conditions---and $\partial \mathcal{D}$ is a subset  of the domain boundary.   Consider a simple feed-forward fully-connected neural network with nonlinear activation functions, where the approximate solution $\tilde{u}(\textbf{x})$ can be further expressed as:
\begin{equation}\label{eqn:pde_gnn}
  \tilde{u}(\textbf{x}, \theta) = \textbf{W}_n\{\phi_{n-1} \circ \phi_{n-2} \circ \cdots \circ \phi_1 \circ \phi_E\}(\textbf{x})+\textbf{b}_n,
\end{equation} 
where $n$ represents the number of neural network (NN) layers, $\phi_l(\textbf{x}_i) = \sigma(\textbf{W}_l\textbf{x}_i+\textbf{b}_l)$ represents the $l^{th}$ layer of the NN,  $\textbf{W}_l \in \mathbb{R}^{d_l\times d_{l-1}}$ and $\textbf{b}_l \in \mathbb{R}^{d_l}$ are the weight and bias of the $l^{th}$ layer, $\phi_E$ is an input encoding layer, $\theta = \{\textbf{W}_1, \textbf{b}_1, \cdots, \textbf{W}_n, \textbf{b}_n\}$ is the set of the trainable parameters of the network, and $\sigma$ is the nonlinear activation function. Given the approximated solution $\tilde{u}(\textbf{x}, \theta)$, we define the following residuals for both PDE equation and boundary (initial) condition constraints:
\begin{equation}\label{eqn:pde_residual}
\begin{aligned}
  &r_{\mathcal{F}}^{(i)}(\textbf{x}, \tilde{u}(\textbf{x}, \theta)) = \mathcal{F}_i[\tilde{u}](\textbf{x})-f_i(\textbf{x}) \\
  &r_{\mathcal{C}}^{(j)}(\textbf{x}, \tilde{u}(\textbf{x}, \theta)) = \mathcal{C}_j[\tilde{u}](\textbf{x})-g_j(\textbf{x}).
\end{aligned}
\end{equation} 

  In order to train the NN model,  residuals can be encoded into a constructed loss function such that $\tilde{u}(x, \theta)$ can gradually approach the true solution $u(x)$ through iteratively optimizing the network parameters $\theta$. The loss function can be represented in the following form:
\begin{equation}\label{eqn:pde_loss}
\begin{aligned}
  \mathcal{L}(\theta) &= \sum_{i=1}^{N_{\mathcal{F}}}{\int_{\mathcal{D}}{w_{\mathcal{F}}^{(i)}(\textbf{x})\|r_{\mathcal{F}}^{(i)}(\textbf{x}, \tilde{u}(\textbf{x}, \theta))\|_p} \,d\textbf{x}}\\ 
  &+ \sum_{j=1}^{N_{\mathcal{C}}}{\int_{\partial \mathcal{D}}{w_{\mathcal{C}}^{(j)}(\textbf{x})\|r_{\mathcal{C}}^{(j)}(\textbf{x}, \tilde{u}(\textbf{x}, \theta))\|_p} \,d\textbf{x}},
  \end{aligned}
\end{equation} 
where $w_{\mathcal{F}}^{(i)}$ and $w_{\mathcal{C}}^{(j)}$ are the weight functions to scale the corresponding loss terms. The model may be trained by stochastic gradient descent such that the model parameters are updated by:
\begin{equation}\label{eqn:sgd_update_rule}
\begin{aligned}
  \theta^{(t+1)} = \theta^{(t)}-\alpha^{(t)}\nabla_\theta\mathcal{L}(\theta^{(t)})
  \end{aligned}
\end{equation} 
\subsection{Importance Sampling  for Training PINNs}
\paragraph{Current State of the Art.} Prior works have explored IS based on the gradient of the loss with respect to the model parameters $\theta$ \cite{Nvidia-IS, katharopoulos2019samplesIS}: 
\begin{lemma}\label{lemma:proportional_update}
Let $H^{(t)} = \nabla_\theta\mathcal{L}(\theta^{(t)})$ the gradient in Equation \ref{eqn:sgd_update_rule}. The convergence of SGD can be accelerated by sampling the input variables from a distribution $P$ that minimizes 
$\mathbf{Trace}(\mathbb{V}_P[H^{(t)}])$, which is accomplished when the probability $P_{x_i}^{(t)}$ of sampling any given point ${x_i}$ at iteration $i$ is 
\begin{equation}\label{eqn:proportional_update}
\begin{aligned}
    P_{x_i}^{(t)} \propto \|H^{(t)}\|_2
\end{aligned}
\end{equation}
\end{lemma}
Additionally, it is shown in \cite{2norm_bounded} that $\|H^{(t)}\|_2$ at a single sample is bounded by a linear transformation of the loss at that sample, and a sampling distribution:
\begin{equation}\label{eqn:proportional_update_loss}
\begin{aligned}
    P_{x_i}^{(t)} \propto \mathcal{L}(\theta^{(t)})
\end{aligned}
\end{equation}
is consistent with Equation \ref{eqn:proportional_update} \cite{Nvidia-IS}. 

\paragraph{Limitations of existing IS methods.} The prior IS method works effectively for training non-parameterized PINNs but may fail for training parameterized PINNs. The sampling probability in (\ref{eqn:proportional_update}) and (\ref{eqn:proportional_update_loss}) only considers the gradient with respect to the neural network parameters ($\theta$) but ignores the impact due to varying input parameters, such as the geometric variations in point clouds. Our experiments using the prior IS method for training parameterized PINNs shows poor generalization capability as discussed in Section \ref{sec:results-parametric}.

\section{The Proposed SGM-PINN  Framework}\label{sec:overview} 
\begin{figure}[!htb]
\begin{center}
\includegraphics[width=.98\linewidth]{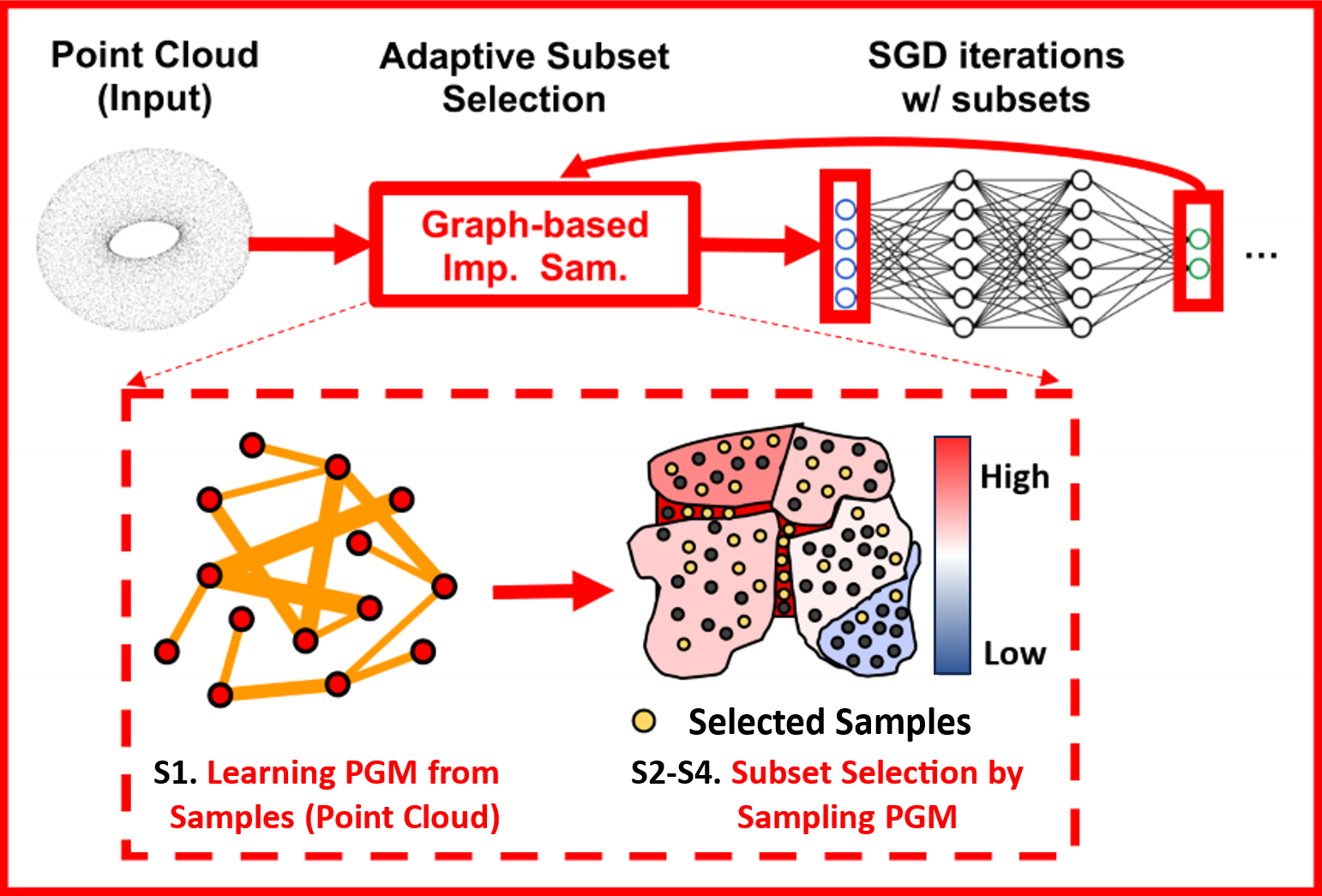}
\caption{The SGM-PINN framework for training PINNs.}\label{fig:overview}
\end{center}
\end{figure}
\subsection{Overview}
\paragraph{The proposed SGM-PINN framework} In this work, we introduce a novel graph-based IS framework, SGM-PINN, for improving the speed and solution quality of the PINN training process. As shown in Figure \ref{fig:overview}, SGM-PINN aims to improve the efficiency of each training step by \textbf{(1)} forming more compact epochs on the fly, and \textbf{(2)} selecting more samples from within clusters that have been scored with high importance. The process consists of four key components: \textbf{(S1)} PGM estimation of the point cloud, \textbf{(S2)} construction of conditionally-independent clusters of nodes via LRD decomposition, \textbf{(S3)} a stability estimation scheme (for parameterized PINNs),  to incorporate additional gradient information with respect to varying input parameters, and \textbf{(S4)} batch ranking and selection for SGD iterations. The proposed sampling strategy allows focusing on the most critical and representative data samples during each training iteration, which immediately improves the overall training process.

\paragraph{Advantages of our approach} There are several potential benefits of the proposed approach: \textbf{(1)} the mini-batch size for the gradient descent algorithm can be reduced, allowing for faster iterations without sacrificing solution quality; \textbf{(2)} the overall dataset size can be reduced, allowing the same models to be run on fewer GPUs or hardware with less available memory; \textbf{(3)} a small subset of samples can be used for estimating the importance scores of the dataset, decreasing the overhead compared to other IS techniques which require updated gradient calculations for every sample; and \textbf{(4)} the reduced overhead achieved by using the PGM to estimate the score for multiple samples can be easily extended to a variety of IS metrics, or combined with new metrics designed to address additional shortcomings in PINN-based solvers, as discussed in Sections \ref{sec:sampling-batches} and \ref{sec:results-parametric}.
\subsection{ Constructing PGMs from Point Cloud (S1) }\label{sec:clustering-PGM}
A probabilistic graphical model (PGM) expresses probabilistic relationships between variables via links between nodes representing the variables \cite{Bishop_2023}. In traditional PDE solvers, discretization methods operate on the principle that the solution at any given point is influenced by its immediate neighbors, with interactions between adjacent mesh or grid points playing a crucial role in problems such as  FEA. In the PINN examples discussed in this paper, a randomized point cloud of $N$ collocation points with $M$ input features $X= \mathbb{R}^{N\times M}$ is generated prior to training as described in \citep{raissi2019physics}.

With the assumption that nearby points will be highly correlated in a relatively dense initial point cloud, we can construct a k-nearest-neighbor (kNN) graph using the low-dimensional spatial coordinates (x,y,z) of each point in the dataset to represent a conditional relationship between points inversely proportional to their distance. This graph, created using an existing highly-efficient kNN algorithm \citep{malkov2018efficient}, serves as an undirected PGM for the input data $X$. At later stages in training this model can be re-built in parallel while incorporating additional features from the output, such as flow or temperature. 
\subsection{Node Clustering by LRD Decomposition (S2)}\label{sec:clustering-LRD}

Node clustering is achieved via a low-resistance-diameter (LRD) decomposition method that allows partitioning any given PGM into multiple node (collocation point) clusters with bounded resistance diameter \footnote{The maximum effective-resistance distances between any two nodes within the same graph cluster will not exceed a given threshold.}. Since the resistance distance between two nodes on a PGM encodes the conditional dependence between the two corresponding data samples, LRD will guarantee that only the most similar data samples will be grouped  into the same cluster, which thereby allows SGM-PINN to select highly-representative data samples during the training of PINNs. 

\begin{definition} \label{def:ER}
  The effective resistance between nodes   $(p, q) \in |V|$ is defined as
\begin{equation}\label{eq:eff_resist0}
    R^{eff}_{p,q} = e_{p,q}^\top L^\dagger e_{p,q}= \sum\limits_{i= 2}^{|V|} \frac{(u_i^\top e_{p,q})^2}{u_i^\top L u_i},
\end{equation}
where $L^{\dagger}$ denotes the Moore-Penrose pseudo-inverse of the graph Laplacian matrix  $L$,   $u_{i} \in \mathbb{R}^{|V|}$ for $i=1,...,|V|$ denote the  unit-length, mutually-orthogonal  eigenvectors corresponding to  Laplacian eigenvalues $\lambda_i$ for $i=1,...,|V|$, and     
${e_{p}} \in \mathbb{R}^{|V|}$ denotes the standard basis vector with all zero entries except for the $p$-th entry being $1$. ${e_{p,q}}=e_p-e_q$.
\end{definition}
In \citep{alev2018graph} it is proven possible to decompose a simple graph  into multiple node clusters of effective-resistance diameter at most the inverse of the average node degree (up to constant losses) by removing only a constant fraction of edges. This is accomplished without significantly impacting the graph conductance (keeping the global structure of the graph intact). 

Directly computing effective resistances according to Definition \ref{def:ER} is   practically infeasible for large PGMs. However, as the proposed LRD method only requires approximate estimation of edge effective resistances, we use a highly-scalable (linear time) algorithm by exploiting a Krylov subspace approach, discussed in detail in \cite{aghdaei2022hyperef}.

Since the proposed decomposition method has a nearly-linear complexity, it can be efficiently applied to   large-scale PGMs (graphs) with millions of data samples (nodes). For additional performance, we also decompose the dataset into grids and perform S1 and S2 in independent sub-processes while training continues either with uniform sampling, or a previously calculated distribution. Speedup is roughly linear with the number of available threads.

\subsection{Stability Score for Parameterized PINNs (S3)}\label{sec:SPADE}

We use a black-box spectral method for assessing the robustness (stability) of ML models and data, relying on graph-based manifold representations~\citep{cheng:icml21}, referred to here as the Inverse Stability Rating (ISR). ISR introduces the concept of Distance Mapping Distortion (DMD), denoted as \(\gamma^F(p,q)\) for a node pair \((p, q)\) transformed by a function \(Y=F(X)\), $F(X)$ in this case being the NN $\tilde{u}(\textbf{x}, \theta)$. It is defined as the ratio of distances between nodes \(p\) and \(q\) on the output and input graphs, $\gamma^F(p,q) \triangleq \frac{d_Y(p,q)}{d_X(p,q)}$, where \(d_X(p,q)\) and \(d_Y(p,q)\) represent the distances between nodes \(p\) and \(q\) on the input and output graphs. Intuitively, the DMD can be employed to estimate the change in distance on the output graph (manifold) due to a perturbation on the input graph (manifold), with the largest $\gamma_{max}^F$ representing the fastest change in output with respect to input.
\begin{lemma}\label{lemma:global_lipschitz}
The ISR is an upper bound of the best Lipschitz constant $K^*$ on F(X)~\citep{cheng:icml21}. 
\begin{equation}\label{eqn:SPADE_K}
\begin{aligned}
    \textbf{ISR}^F \stackrel{\text{def}}{=} \lambda_{max}(L_Y^+L_X) \geq K^* \geq \gamma_{max}^F
\end{aligned}
\end{equation}
\end{lemma}
\begin{lemma}\label{lemma:local_lipschitz}
 ISR defines the edge score between two nodes $p,q$, $\textbf{ISR}^F(p,q)$ where $V_r\overset{\mathrm{def}}{=}\left[ {v_1}{\sqrt {\lambda_1}},...,  {v_r}{\sqrt {\lambda_r }}\right]$, and $\lambda_r$, $v_r$ represent the first $r$ largest eigenvalues and corresponding eigenvectors of $L_Y^+L_X$ as~\citep{cheng:icml21}:
 \begin{equation}\label{eqn:SPADE_EDGE}
 \textbf{ISR}^F(p,q)\stackrel{\text{def}}{=} |V_r^\top e_{p,q}\|_2^2 \propto \left(\gamma^F(p,q)\right)^3
 \end{equation}

Then, individual sample stability can be quantified as the average edge score among all $q_i$ of $p$, the set of which is denoted by $\mathbb{N}_X(p)\in V$~\citep{cheng:icml21}:
\begin{equation}\label{eqn:SPADE_NODE}
\textbf{ISR}^F(p)\overset{\mathrm{def}}{=}\frac{1}{|\mathbb{N}_X(p)|}\sum_{q_i\in \mathbb{N}_X(p)}^{}{|V_r^\top e_{p,q_i}\|_2^2} 
\end{equation}
\end{lemma}

 Each edge score acts as a surrogate for the directional derivative between two nodes on $F(X)$\cite{cheng:icml21}, which in our application are the NN losses, such that:
\begin{equation}\label{eqn:SPADE_GRADIENT}
\textbf{ISR}^F(x_i)\geq ||\nabla_{x_i}\mathcal{L}(\theta)||_2
\end{equation}

In \cite{Nvidia-IS} additional computational efficiency is achieved by partitioning nodes among a much smaller set of random "seeds"\cite{Nvidia-IS}, updating the loss value at each seed to calculate $\mathcal{L}(\theta^{(t)})$ and assigning that value to nearby samples in a piece-wise fashion before calculating $P_{x_i}^{(t)}$ for each sample. In our method, we also use nearby neighbors to adjust the probability of sampling points we did not directly update the loss for, but we can add additional weight to clusters with high ISR. As it represents quickly changing local losses, it implies that the loss estimate for that cluster may be poor, and potentially important points may be ignored.

\subsection{Batch Ranking and Selection (S4)}\label{sec:sampling-batches}
Only $r$ points in each cluster of samples are updated with the latest losses to rank that cluster relative to others, allowing a reduced number of passes through the NN to update $P$.
Sampling is maintained with the principle in Equations \ref{eqn:proportional_update},\ref{eqn:proportional_update_loss} by sampling more from clusters with losses higher than other clusters, as described in Algorithm \ref{alg:pinn}.
There is a floor of 1 sample per cluster when determining an epoch, so no cluster is entirely ignored in any epoch. This mitigates the potential for failing to refine or 'forgetting' the solution \cite{daw2023mitigatingR3} in areas with relatively low scores as training continues.
For parametric examples the ISR discussed in S3 is added as an additional term to the cluster score prior to ranking, using the same subset of samples as---and normalized with---the other PDE losses. S3 is also performed on a background thread, so the only additional overhead on GPU resources and overall wall time is the $r \times N$ loss calculations every $\tau_e$ iterations.

\subsection{Algorithm Flow and Complexity}\label{sec:alg-flow}
\begin{algorithm}
\small { \caption{The algorithm flow of SGM-PINN}\label{alg:pinn}}
\begin{flushleft}
\textbf{Input:} Sample matrix $X = \mathbb{R}^{N\times M}$ of $N$ samples with $M$ selected features, the ratio $r$  of points in each cluster to sample for estimating loss. $\tau_e$ is the number of times to repeat the epoch. 

\textbf{Output:} An epoch of mini-batches for training. 
\end{flushleft}
  \algsetup{indent=1em, linenosize=\small} \algsetup{indent=1em}
    \begin{algorithmic}[1]
    \STATE From $X$, create a kNN-graph $G$
    \STATE Use the LRD Algorithm to  split $G$ into $n_c$ clusters of similar samples.
    \STATE $S \gets$ an array of the sizes of each cluster.
    \WHILE{$step\_count$ $<$ $step\_target$}
    \STATE $S^* \gets$ {$r\cdot S_i$} points from each cluster in S 
    \STATE Calculate the losses for $S^*$
    \STATE From S*, apply the ISR algorithm.
    \STATE $L \gets$ combined losses and ISR for each cluster
    \STATE Map $L$ to a range of proportional sampling ratios $P$
    \STATE Create an epoch with of {$P_i\cdot S_i$} samples from each cluster
    \WHILE{$step\_count\%\tau_e \neq 0$}
      \STATE Shuffle and return the epoch
    \ENDWHILE
    \IF{$\tau_G$ has passed}
      \STATE Start 1,2,3 in a background task
    \ELSIF{$S_{new}$ ready}
      \STATE $S \gets$ $S_{new}$
    \ENDIF
    \ENDWHILE
    \end{algorithmic}
\end{algorithm}

Algorithm \ref{alg:pinn} shows the key steps in SGM-PINN. Step 1 kNN construction using the HNSW algorithm \cite{malkov2018efficient} has a complexity of $O(N\log (N))$, where $N$ is the number of nodes. Step 2 LRD is a nearly-linear time algorithm with a runtime proportional to the number of edges the graph $G$, which is determined by $N \times k$, $O(kN)$. Sampling losses for importance scores are an additional overhead cost, with the number of extra forward passes determined by $\tau_e \times r \times N$ for relatively frequent loss updates, as well as $\tau_G \times N$ for infrequent $G$ updates. ISR has a complexity $O(N\log(N))$.
This gives an overall complexity of $O(N\log (N)) + O(kN\tau_e + N\tau_G)$.

\section{Experimental Results}\label{sec:results}
 
We implement the  proposed SGM-PINN framework on Nvidia's Modulus v22.09 (previously on SIMNET v21.06) platform \citep{hennigh2021nvidia}. We conduct extensive experiments for evaluating the performance and efficiency of the proposed SGM-PINN framework. All tests were run on a Xeon Gold 6244 CPU @ 3.60GHz, 1.5TB of 2933 MHz available system memory, and 1 Tesla V100 32GB GPU. Examples are shipped with Modulus and use the default settings where applicable. An implementation is available online at \url{https://github.com/Feng-Research}.

Modulus also includes an importance sampling implementation based on \citep{Nvidia-IS} (labeled MIS here) which we benchmark against our method. This method assigns a sampling probability based on the 2-norm of the velocity ($u$,$v$) derivatives. For an even comparison we reduce how often the dataset is updated to match $\tau_e$. By default MIS re-calculates sample probabilities every epoch and may perform slower than the baseline random sampling due to loss updates not contributing to training. We additionally only apply MIS to the sampling of interior points, as SGM-PINN has not yet been implemented for the boundary or initial conditions.

The two example problems presented in this paper are simulations related to computational fluid dynamics (CFD). The first, Lid Driven Cavity (LDC) \citep{nvidia-man} is a well-studied benchmark example, with the addition of a zero-equation turbulence model. Outputs measured against OpenFOAM \cite{openfoam} validation data are $u$ (x-velocity), $v$ (y-velocity), and $\nu$ (kinematic viscosity). Further details are in Section \ref{sec:results-LDC}
The second, annular ring (AR)\citep{nvidia-man} is another 2D laminar flow example that considers the flow from an inlet to an outlet through a symmetrical annular ring with a parameterized inner radius $r_i$. Outputs are $u$, $v$, and $p$ (pressure), with validation results available at $r_i$=1.0, $r_i$=0.875, and $r_i$=0.75. All NNs use a fully connected architecture with width 512, depth 6, and SiLU\citep{silu} activation functions. Data presented is an average of 5 runs for each example.

\subsection{LDC, Non-Parametric, without S3}\label{sec:results-LDC}
\paragraph{Experimental Setup.} LDC with zero-eq turbulence, the Reynolds number $Re=1000$, and the top wall is moving at $1 m/s$. $U_{4000}$ is the 'baseline' example with uniform random sampling, batch size $\beta=4000$, and number of samples $N=16M$. The remaining methods $U_{500}$, $MIS_{500}$, $SGM_{500}$ have $\beta=500$ and $N=8M$. $MIS_{500}$ and $SGM_{500}$ re-calculate sample  scores ($\tau_e$) every 7k iterations. $SGM_{500}$ uses $r=15\%$ samples per cluster and fully recalculates clusters ($\tau_G$) every 25K iterations with kNN size $k=30$ and LRD level $\mathbb{L}=10$. 

\begin{table}[]
\caption{Minimum Validation Errors and Time to Achieve for LDC\_zeroEq. $T(M_\beta\_j)$ is the time taken by a sampling method in the top row to achieve $Min(j)$ of $M_\beta$. Blanks are left where that value was not achieved. The best and second-best results are bolded and italicized, respectively. The diagonal (time each took to get to its own best value) is underlined.}
\label{tab:ldc}
\begin{tabular}{lllll}
\toprule
Label         & $U_{500}$   & $U_{4000}$  & $MIS_{500}$            & $SGM_{500}$         \\
   & &  &   & (ours)            \\
\midrule
$Min(u)$            & 0.0879 & 0.0480           & \textit{0.0431}   & \textbf{0.0412} \\
$Min(v)$            & 0.1169 & \textit{0.0589}  & 0.0623            & \textbf{0.0573} \\
$Min(nu)$           & 0.2173 & 0.1788           & \textit{0.1735}   & \textbf{0.1595} \\
$T(U_{4000}\_u)$    &    -    & \underline{32.55}& \textit{16.53}    & \textbf{9.48} \\
$T(MIS_{500}\_u)$   &    -    &         -         & \textit{\underline{24.19}} & \textbf{11.26} \\
$T(SGM_{500}\_u)$   &    -    &          -        &                   -&\underline{\textbf{16.91}}\\
$T(U_{4000}\_v)$    &    -    &\textit{\underline{32.61}} &          -         & \textbf{11.45} \\
$T(MIS_{500}\_v)$   &   -     & \textit{30.23}            &\underline{35.29}  & \textbf{10.25} \\
$T(SGM_{500}\_v))$  &    -    &         -         &          -         & \underline{\textbf{18.18}} \\
\bottomrule
\end{tabular}
\end{table}

\begin{figure}
\centering
    \graphicspath{{./results/}{./results_new}{./}}
 	\includegraphics[width=0.875\linewidth]{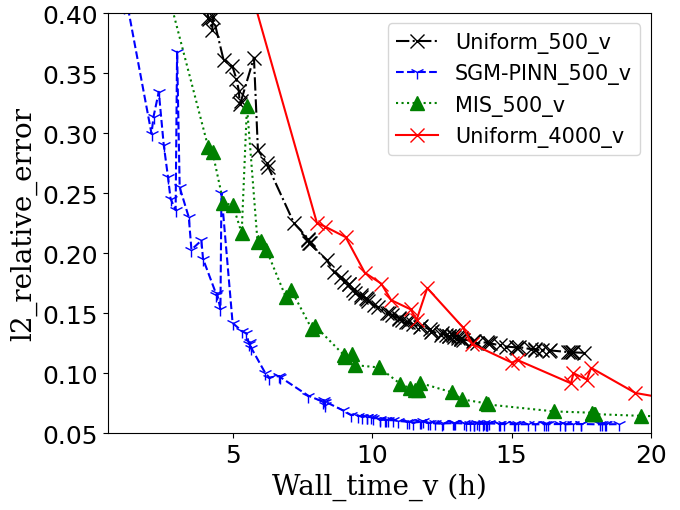}
	\caption{Solution error by wall time (lower) for $v$ in the LDC example. $Uniform_{500}$, $SGM$-$PINN_{500}$, and $MIS_{500}$ have batch sizes of 500 and a total of 500,000 collocation points. The baseline $Uniform_{4000}$ has a batch size of 4000 and 4M total collocation points.}\label{fig:ldc}
\end{figure}

As shown in Figure \ref{fig:ldc} and Table \ref{tab:ldc}, SGM-PINN improves both runtime and accuracy for the LDC example. 
Targeting the same accuracy as the baseline, SGM-PINN achieves a runtime improvement of $3.43\times$ in $u$ and a $2.85\times$ in $v$. The best result is achieved $1.79\times$ faster than the baseline's, with a $14\%$, $3\%$, and $11\%$ reduction in relative error in $u$, $v$, and $nu$ respectively compared to the baseline. The runtime improvement in $u$ and $v$ compared to the built-in importance sampling technique is $2.15\times$ and $3.44\times$. Reducing the batch size to 500 and taking the total sample count to only 500,000, SGM-PINN achieves at least $2.5\times$ speedup while reaching the same accuracy across all variables compared to the baseline. It completes 2.5 million iterations in 19 hours compared to 1 million in 25 hours by $Uniform_{4000}$. Without either MIS or SGM-PINN, reducing the batch and dataset size with uniform sampling cuts the final solution quality in half with no benefit to convergence speed. 

\subsection{Parameterized Annular Ring, with S3} \label{sec:results-parametric}
\begin{figure}[!htb] 
\centering
    \graphicspath{{./results/}{./results_new}{./}}
 	\includegraphics[width=0.8995\linewidth]{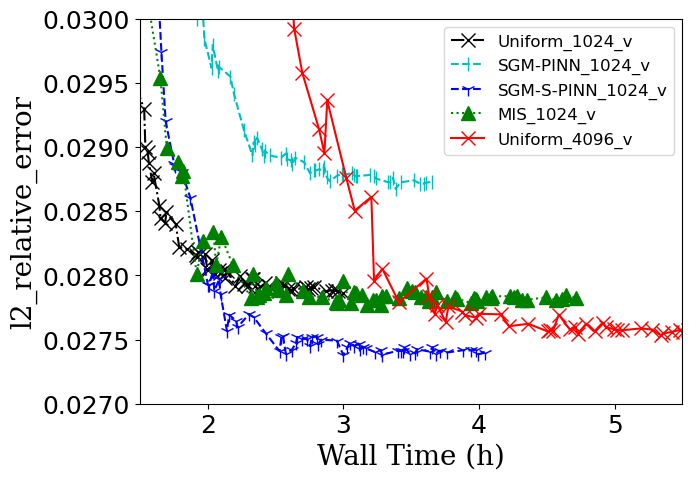}
	\caption{Solution errors of $v$ for parameterized PINN for the AR example compared to the OpenFOAM validation data averaged at $r_i$=1.0,0.88, and 0.75, respectively.}\label{fig:main_ar_v}
\end{figure}

\begin{figure}[!htb] 
\centering
    \graphicspath{{./results/}{./results_new}{./}}
 	\includegraphics[width=0.85\linewidth]{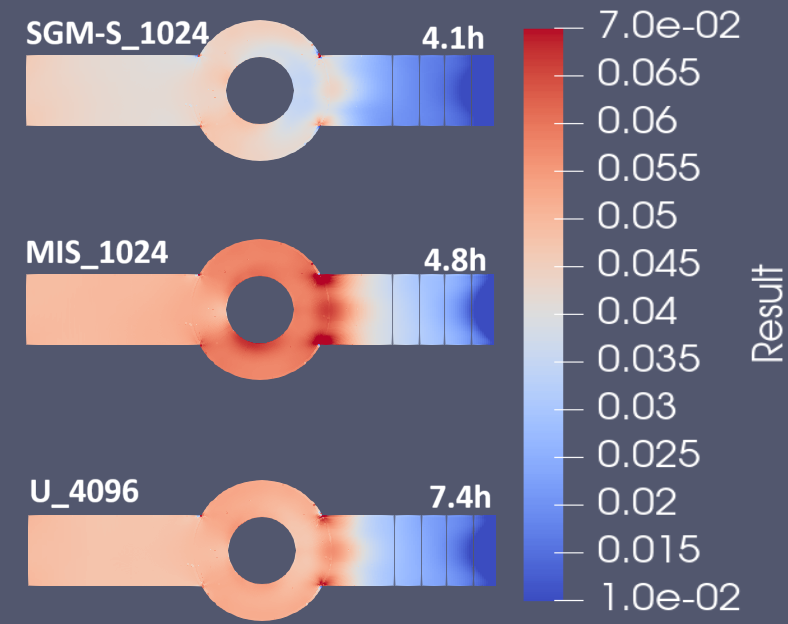}
	\caption{Visualized absolute errors for $p$ at $r_i$=1.0} \label{fig:main_ldc}
\end{figure}

\paragraph{Experimental Setup} Physical parameters are inlet velocity of 1.5m/s, channel width of 2m, and viscosity $\nu=0.1$ with a parameterized interior radius of $r = [0.75,1.1]$. $U_{4096}$ is the 'baseline' example with uniform random sampling, batch size $\beta=4096$, and number of samples $N=16M$. The remaining methods $U_{1024}$, $MIS_{1024}$, $SGM_{1024}$, and $SGM$-$S_{1024}$ have $\beta=1024$ and $N=8M$. ($\tau_e=7$K) for $SGM$ and $MIS$. For both $SGM$ methods $k=7$, $\mathbb{L}=6$, $r=15\%$ and $\tau_G=60$K.
A key application for PINN solvers is the ability to solve a problem across parameterized geometries \citep{hennigh2021nvidia}\citep{PINN_parametric_ability}. Both SGM-PINN and MIS have trouble with the parameterized example compared to uniform importance sampling, as shown in Figure \ref{fig:main_ar_v} and summary Table \ref{tab:ar}. It is observed that for parameterized training SGM alone decreases performance (SGM-PINN in Figure \ref{fig:main_ar_v}). MIS additionally decreases performance, although to a lesser degree. Including the stability metric (SGM-S-PINN) in our method allows us to maintain accuracy in $u$ and $v$ relative to the baseline uniform sampling, while improving $p$ by 12\% (from 0.132 to 0.116) despite training for half the time. The average error for $v$ during training is provided in Figure \ref{fig:main_ar_v}. Figure \ref{fig:main_ldc} visualizes the solution errors for $p$ at 350K iterations for all methods, showing our method has the lowest error compared to validation data while taking the least time.
\begin{table}[!htb]
\centering
\caption{Results for Annular Ring Parameterized, averaged. The validation value for $p$ does not monotonically decrease to a consistent value during training---it reaches a minimum well before $u$ and $v$ are trained and then trends upward before levelling out. Since $v$ takes longest to converge to a minimum, the value for $p$ is given then.}\label{tab:ar}
\begin{tabular}{lllll}
\toprule
Label             & $U_{1024}$    & $U_{4096}$    & $MIS_{1024}$  & $SGM$-$S_{1024}$ \\
\midrule
$Min(u)$              & 0.0289                      & \textit{0.0285} & 0.0294        & \textbf{0.0278}  \\
$Min(v)$              & 0.0279                      & \textit{0.0275} & 0.0278       & \textbf{0.0274}  \\
$p$ at $Min(v)$       & \textit{0.123}              & 0.132           & 0.137        & \textbf{0.116}    \\
$T(U_{1024}\_u)$      & \textit{\underline{2.20}}   & 3.23            &        -      & \textbf{2.00}    \\
$T(U_{4000}\_u)$      &                     -    &\textit{\underline{3.63}}&      -        & \textbf{2.00}    \\
$T(MIS_{1024}\_u)$    & \textbf{1.65}	            & 3.02            &\underline{2.32}& \textit{2.00}    \\
$T(SGM$-$S_{500}\_u)$ &      -          &         -        &       -       &\underline{\textbf{2.14}}\\
$T(U_{1024}\_v)$      &\underline{3.00}& 3.41            & \textit{2.32}         & \textbf{2.09}    \\
$T(U_{4000}\_v)$      &      -          &\textit{\underline{6.19}} &       -       & \textbf{2.53}    \\
$T(MIS_{1024}\_v)$    &      -          & 3.67            &\textit{\underline{3.28}}& \textbf{2.14}    \\
$T(SGM$-$S_{500}\_v)$ &       -         &     -            &      -        &\underline{\textbf{3.00}}    \\
\bottomrule
\end{tabular}
\end{table}
\section{Conclusion}\label{sec:conclusion}
In this work, we introduce a graph-based sampling framework for speeding up the training of PINNs. SGM-PINN allows the importance score of multiple samples to be estimated via selection of highly-correlated clusters within the point cloud by leveraging PGMs. We additionally demonstrate the inclusion of a stability score in calculating importance for improving parameterized training. Our experiments show SGM-PINN leads to a $2\times$-$3\times$ runtime improvement for training PINNs related to CFD problems.  
Future work will focus on further reducing the overhead for implementing SGM-PINN and including importance sampling on BCs. More complex examples can also be sensitive to the hyper-parameters $k$ and $\mathbb{L}$, as is the performance overhead. Automatic tuning of these parameters would be preferred. Additionally, much larger PDE problems from more varied domains need to be tested.

\section{Acknowledgment}
This work is supported in part by the National Science Foundation under Grants CCF-2212370, CCF-2205572, and CCF-2021309.

\bibliographystyle{ACM-Reference-Format}
\bibliography{i,f,j,new}

\appendix

\end{document}